# MATra: A Multilingual Attentive Transliteration System for Indian Scripts

Yash Raj[1] and Bhavesh Laddagiri[2]

[1]Chandigarh University
[2]SRM Institute of Science and Technology

*Abstract*—Transliteration is a task in the domain of NLP where the output word is a similar-sounding word written using the letters of any foreign language (for example, the English word 'LEAGUE' is written as 'लीग' in Hindi). Today this system has been developed for several language pairs that involve English as either the source or target word and deployed in several places like Google Translate and chatbots. However, there is very little research done in the field of Indic languages transliterated to other Indic languages. This paper demonstrates a multilingual model based on transformers (with some modifications) that can give noticeably higher performance and accuracy than all existing models in this domain and get much better results than state-of-the-art models. This paper shows a model that can perform transliteration between any pair among the following five languages - English, Hindi, Bengali, Kannada and Tamil. It is applicable in scenarios where language is a barrier to communication in any written task.

The model beats the state-of-the-art (for all pairs among the five mentioned languages - English, Hindi, Bengali, Kannada, and Tamil) and achieves a top-1 accuracy score of 80.7%, about 29.5% higher than the best current results. Furthermore, the model achieves 93.5% in terms of Phonetic Accuracy (transliteration is primarily a phonetic/sound based task).

*Index Terms*—Transliteration, Multilingual Transliteration, Indic Transliteration, Transformers, Natural Language Processing, NLP

## I. INTRODUCTION

TRANSLITERATION is a relatively old field of study [1], where the letters of a different language are used to write words of the source language by matching the sound of the source language word. For example, the word 'LEAGUE' in English will be written as 'लीग' in Hindi. Anyone who does not know how to read English properly, while they are fluent in reading Hindi, this task can assist them in reading and speaking such English words by reading the corresponding Hindi transliterated word.

Several models and other approaches have been built that generally involve the input word of any language that is transliterated to English. In the industry, the most prominent use of such models is done by Google in their Google Translate website, where the input is the word/sentence from the native language and translated word is in the chosen language. But along with the translated word, Google Translate gives the transliterated form of native word/sentence in English to assist in pronouncing such words/sentences of foreign languages.

Several systems have been built in this domain that primarily depend on rule-based approaches, with the recent addition of ML-based models to increase the accuracy of this task. However, most of them are done for the task of transliteration from other languages to English. There are very few models that have been developed for the task of transliteration from English to other languages. Furthermore, minimal research is done for transliteration of Indian languages to other Indian languages [2]. India is a diverse country with 121 different languages spoken, and it is necessary to bridge this language barrier. The model presented in this paper will help solve the language divide as it can help anyone read and speak new languages with the help of languages they can already read and speak. The model presented in this paper can transliterate text between any of the pairs among the five languages - English, Hindi, Bengali, Kannada and Tamil. These are the five most common languages spoken across the country. Translating text between these languages would help the majority of the population who face any issues reading the other language.

Transliteration [3] is a task defined in the domain of NLP (text and written words) where the output contains the letters of a foreign language while the sounds and phonetics are from the native language. The output word is essentially based on the native language as it conveys the same meaning, just written with the letters of the foreign language. The term 'transliteration' combines two root words - trans (meaning different) and liter (meaning letters). Thus, the output word that contains letters from the foreign language will not hold any meaning in the foreign language, i.e., a word might never exist in the vocabulary of the output language. Taking the language of Hindi as an example, we can transliterate the Hindi word 'पाठशाला' as 'pathshala' in English. Naturally, this word does not exist in the English vocabulary.

This task is similar to translation [4] where the words and letters are always part of a foreign language. The difference is that the output here does not hold any meaning in the foreign language. In contrast, in translation, the output words are part of the foreign language, and those words hold the meaning intended to be conveyed.

### A. Challenge in Transliteration

In the task of translation, the output sentence can be constructed with only a fixed set of words and the output words are always ordered in such a way that they make sense. They must convey meaning based on the words preceding and following it. For example, the word 'class' in English can only be translated as 'पाठशाला' in Hindi and no other word or set of letters would work, or else the meaning would change completely.

On the other hand, any word like 'pathshala' can be back-transliterated [5] as 'पाठशाला' or 'पाठषाला'. Both these are perfectly correct as the phonetic sounds of both words are identical (see Fig 1).

Some other ambiguous examples for transliteration, taken directly from the dataset are -

1) 'किन' can be transliterated as 'QIN' or 'KIN'
2) 'लीग' can be transliterated as 'LEAGUE' or 'LEEG' or 'LIIG'
3) 'फीका' can be transliterated as 'FEEKA' or 'PHEEKA' or 'PHIKA' or 'FIKA'

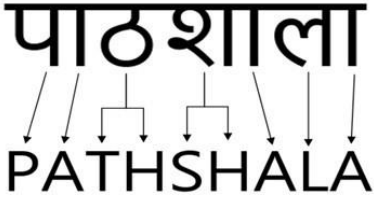

Fig. 1: Example of Transliteration from Hindi to English. Arrows show which letter(s) are mapped to which exact letter(s). However, every example is not so easy to map letter wise.

Due to the availability of multiple output words (as phonetic sounds are precisely similar) for each input word, it is extremely tough to find the exact word. Even the dataset has multiple words for training the model to become robust and capable of handling such predictions. The uncertainty of the correct word makes this task very tough to evaluate. These predictions are perfectly correct for a human, so only human evaluators can assess the model's predictions. If we keep accuracy as the metric and use the dataset as a benchmark, we will have numerous words mispredicted. At the same time, they are phonetically correct and must be labelled correctly. However, the test dataset has only one correct prediction of each word, and our model can handle multiple predictions. For example, for the word 'pathshala', the test dataset gives 'पाठशाला' as the correct prediction. However, if the model predicts 'पाठषाला', it is marked incorrect (but is correct, as discussed above.) Even for human evaluators, it is relative, and some strict evaluators might mark 'पाठषाला' as incorrect (which is the answer given in the test dataset.) Thus, this task requires extensive care while evaluating (Fig 2).

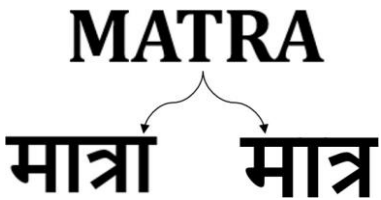

Fig. 2: Transliteration of the model's name from English to Hindi (showing 2 possible correct outputs). Here, both the word can not be mapped letter-wise.

Fig 2 shows the transliteration of our model's name from English to Hindi. Even for such basic words like the model's name, there are 2 possibilities and both are perfectly correct.

## II. RELATED WORK

Indic-to-Indic transliteration is a niche problem statement and there hasn't been active research in this area. Only a few papers in the past few years have made any contributions in this field.

Leveraging Orthographic Similarity for Multilingual Neural Transliteration [6] was based on a method that used orthographic similarities to solve this task. The paper used CNN and LSTMs for the task of multilingual transliteration. This model could also do zero-shot multilingual transliteration in some instances for foreign languages (on new languages). The dataset used in this paper is unreleased.

Brahmi-Net [7] is a network based on statistical techniques. The paper performs multilingual transliteration either directly or through an intermediate language for some of the challenging pairs. They used a massive dataset for training, which wasnot publicly released.

## III. MODEL

### A. Background on Transformers

Previously, NLP had been largely dominated by Recurrent Neural Networks (especially LSTMs [8]) due to the sequential nature of text where the next words or characters in a piece of text depends on the previous words or characters. LSTMs were a popular choice because it was able to process the words sequentially and remember the history as context when making predictions. They were also able to take arbitrary length inputs instead of fixed length inputs as seen in traditional Deep Neural Nets. However, with longer sequences, the performance of the models fell. To combat this, attention mechanism [9] [10] was introduced to focus on the relevant parts of the sentence to make predictions. This gave a major boost in performance to neural machine translation systems. But some fundamental issues in recurrent architectures were still unsolved. These were primarily,

- vanishing and (or) exploding gradients.
- slow training due to the sequential nature of feeding inputs to the model resulting in a low throughput.
- struggles with remembering long sequences of text.

Given, that attention had given a significant boost in performance, the Transformers [11] were introduced in 2017 which purely uses attention only without any recurrence or inductive biases towards the sequential nature of language. Transformers has become the de-facto standard for natural language processing today. The most dominant approach to using transformers is to pre-train on a large corpus of text and then fine-tune on a downstream task [12]. Transformers being a general architecture without any inductive biases has led to it being successfully applied in other domains like computer vision [13], reinforcement learning [14] and speech [15], among others. It is safe to say that transformers have greatly impacted the entire field of machine learning.

The transformers were originally designed in a encoder-decoder structure [16] where the input sequence is encoded to a feature vector on which the decoder is conditioned to auto-regressively [17] generate the target sequence.

*1) Encoder:* The encoder stack is primarily made of two sub-layers, multi-headed self attention and fully-connected layer connected with residual connections [18] around each sub-layer. The input sequence is converted to a vector using a embedding layer and a positional embedding is also added to it before being fed into the encoder.

*2) Decoder:* The decoder stack additionally employs a third sub-layer, apart from the two from the encoder, which performs multi-head attention on the outputs of the encoder. The first self-attention layer in the decoder is also modified to prevent tokens attending to future tokens in the output sequence during training. This done by masking the output sequence with a triangular attention mask such that a token in position $i$ can only depend on positions less than $i$.

**Our model is very similar to the original transformers' architecture with slight changes in the training method**

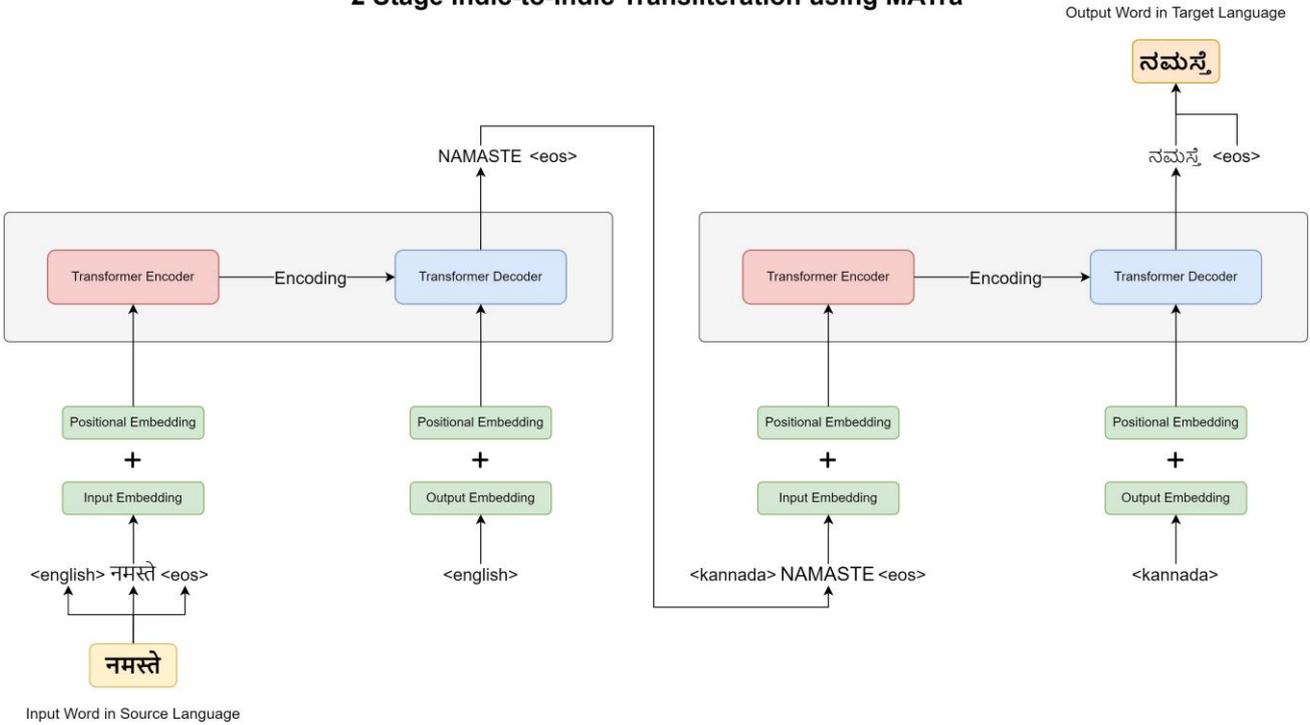

Fig. 3: The pipeline here shows the 2-stage approach to transliterating between Hindi and Kannada using our Multilingual Bi-Directional Model, *MATra*. In stage 1, the Hindi word is prefixed with the `<english>` token and transliterated to English which serves as our intermediate representation. This English transliterated word is then fed back into the model with the target language token i.e., `<kannada>` which produces the final transliteration in our target Indic language.

and using positional embeddings instead of sinusoidal encodings.

### B. Multilingual Training via Special Tokens

Sequence to sequence models for NLP are trained with special tokens as prefix and suffix which indicate whether the sentence is starting or ending. These tokens are especially useful for the decoder as the token indicating the start of the sequence is usually fed as the initial input and as the decoder auto-regressively generates the sequence, it stops when the next token predicted is a token indicating the end of sentence, as shown below:

`<SOS> Some Text Sequence <EOS>`

However, instead of using the usual `<SOS>` token [19] as the input to the decoder we employ special language tokens, like `<hindi>`, `<bengali>`, etc., that guide the decoder in generating the transliterated sequence in the target language, as shown below:

`<hindi>` लीग `<EOS>`

To allow the model to adapt the encodings of the source language with respect to the target language, we use the language tokens as the start token in the input sequence as well. However, in practice we found out that the encodings only differed by 1 percent, compared to encodings of the same input sequence but with different language tokens at the start. Nevertheless, the language tokens serve their primary purpose to the decoder in generating multilingual outputs by simply changing the language in the seed input to the decoder.

### C. Uni-Directional Baselines

The datasets that are available for us to work with have English as either the source or the target. As a baseline we begin with training models that are multilingual at either the source or the target at a time. We call these models uni-directionally multilingual.

*1) Indic to English:* This model is a many-to-one transliteration model with multiple Indic languages as the input (many) and English as the output (one). As the decoder always produces English, the use of language tokens isn't necessary.

*2) English to Indic:* This model is a one-to-many transliteration model with English as the input (one) and multiple Indic languages as the output (many). This model is trained on just the inverse dataset of the Indic to English model. In this case the decoder produces multiple languages and the language token is necessary to guide the model to the desired target language.

To perform multilingual transliteration between any indic language, we can chain these two uni-directional models together. For example, to transliterate between Hindi and Bengali we can use the Indic to English model to get the English representation of the Hindi word and then feed this English representation to the English to Indic model to get the Bengali word as the output.

### D. Our Bi-Directional Model

For Multilingual Transliteration, the chaining of Uni-Directional Models works decently well, however, it still occupies two models in memory and the performance of transformer models increases with size of the dataset. To resolve these two aspects, we finally train a Bi-Directional transliteration model trained with data that contains training examples from both the uni-directional baselines. Hence, it can perform both Indic to English and English to Indic transliteration and as it's a many-to-many task, the language tokens are used as the start token in both, the source sequence and the target sequence. During inference, the same chaining

principle is used as with the uni-directional models as shown in Fig 3

Our model follows the standard architectural configuration as most language models that are out there like BERT [12]. The configuration is as follows:

```
num_encoder_layers = 12
num_decoder_layers = 12
embed_size = 768
heads = 12
batch_size = 32
scheduler = 'linear with warmup'
warmup_steps = 300
epochs = 16
hidden_dim = 3072
max_seq_len = 50
dropout = 0.0
```

IV. EXPERIMENTAL SETUP

Supplementary details about the dataset used, and few outputs are given in the following sections - IV.A and IV.B. Furthermore, the model is deployed as an API and can be accessed from the website, whose details are mentioned in section IV.B.I.

*A. Dataset*

For this task of multilingual transliteration, we used the NEWS 2018 and NEWS 2012 datasets combined. Both the datasets have several transliterated words from English to several other languages. For our task, we are transliterating from the following 5 languages to any of the 5 languages -

1) English

2) Hindi

3) Bengali (Bangla)

4) Tamil

5) Kannada

The datasets to each language are present in the NEWS 2018 and NEWS 2012 as separate XML files which can be parsed to extract each pair. Fig 4 shows the number of words in each file after processing and cleaning each file separately.

Using these datasets themselves, we had to solve the task of multilingual transliteration. Other datasets were present, like the Dakshina dataset (also based on English to Indic format), which had more samples, but it was very impure and, thus, not wholly reliable. Several words in the Dakshina dataset were misspelt, and a few examples were pure translation examples or completely ambiguous and unrelated to transliteration. Furthermore, there were no other current datasets (created in the last 2 to 3 years) that could be used because all of them were impure and had samples that were incorrect. To maintain the purity of our dataset and only keep all correct examples, we chose not to use the Dakshina dataset or other similar recent datasets (which had incorrect samples and were highly inconsistent).

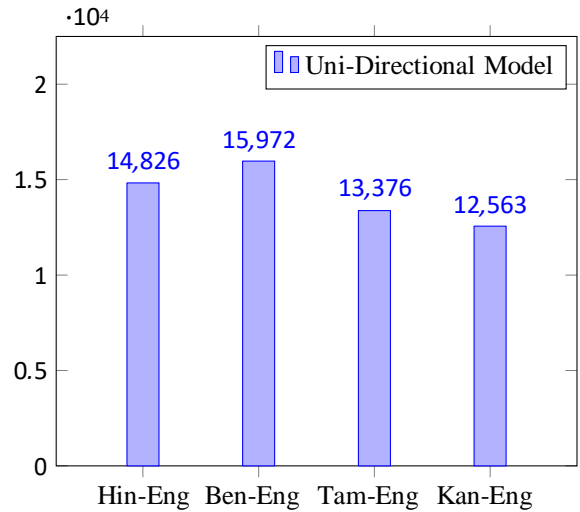

Fig. 4: Shows the number of words in each file (for each language)

*1) Issues in the dataset :* The datasets for English to the respective four languages are present, and upon reversing them, we can obtain the respective languages to English transliterated pairs. Nevertheless, anyone would notice that there is no direct dataset for this task. For example, transliteration from Hindi to Bengali has no dataset available. Creatinga new dataset for each language pair is a monumental task as we do not have the resources to do so.

Furthermore, each English word in all datasets is primarily unique, and there are very few examples where the same English word is present in 2 different datasets. For example, if the word "hello" is present in the English-Hindi dataset, there is a meager chance it will be present in any other dataset. Similarly, most words are unique, so we cannot even take the standard matches from 2 or more datasets and generate the mappings. The only solution we had was to train without a direct dataset, and this is further elaborated in the model section (section III). However, we created a dataset with all the word pairs from each language (independent of each other) and used that for training. The final dataset had pairs from all other languages to English and English to all other languages. This combined dataset finally had 113,474 pairs (113k approximately), and we trained our model on this dataset itself. Though this dataset did not have even one pair from any Indian language to the other Indian language (for example, there were no pairs from Bengali to Kannada, or Hindi to Tamil, and so forth), we were still able to train a model that could do this task from any Indian language to the other Indian language indirectly (for example, transliterate from Bengali to Kannada, or Hindi to Tamil, etc.)

To further add to this issue of no direct dataset, the training of transformers requires massive datasets to scale and attain high performance. Comparing the multilingual models like mBERT [20] and mT5 [21], we can quickly notice that our dataset is tiny as compared to the datasets on which those models were trained. It is a known fact that training transformers require massive datasets in general else the model will not learn much and not generalize well. mBERT is trained on Wikipedia and other datasets, and comparing the dataset size it is evident that we have a tiny dataset compared to the other multilingual models in the industry.

*2) Pre-Processing and cleaning of dataset:* The NEWS 2018 and NEWS 2012 dataset is the purest among all the available datasets, but still is not 100% correct. This dataset was present in the XML format and it had to be parsed to extract the input and output words separately. Following steps were implemented in order to get the final pairs -

1) The XML files were parsed for each dataset and each data point was stored in a list (containing the input word as the first element, and output word as the second element.)

2) Most examples are in the format of 1 input to 1 exact output, but few input words had 2 or more correct outputs. These extra examples arise when 2 similar sounding letters are used. We had to generate 2 or more pairs by repeating the input word depending on the number of outputs words given in the dataset.

3) The task of transliteration needs to be count of words specific, i.e., for 1 input word, there can be only 1 output word, or for a sentence of 10 words, the output sentence must also be exactly 10 words in length. The length of each word can be language specific but count must be exact. There are few pairs in the datasets which had different number of words in the input and output and they can not be used for final training, so we had to eliminate such examples.

4) Few words had letters which was not part of the respective language letters. Such letters had to be removed, or the word had to be completely eliminated, depending on the word.

5) A special language token was added at the end of each pair, as the third element in the list. The use of language token is further elaborated in (section III.B). But this language token was based on the output language (i.e., language in which the output word is written.)

Finally, we had a list containing 113k language pairs followed by the output token. A small example of the English to Hindi and Hindi to English dataset is as follows :
[ किन , QIN , <english> ], [ लीग , LEAGUE, <english> ],......,[ QIN , किन , <hindi> ], [ LEAGUE , लीग , <hindi> ], .....
This is the same format for all other language pairs like Bengali to English and English to Bengali, and similarly for Kannada and Tamil. But the original issue of no direct dataset still remains, and there are no pairs from Hindi to Bengali (or any Indic-to-Indic language).

B. *Outputs from the Model*

Few examples for each language pair are given in Tables I to V. All these examples are chosen at random with no specific criterion. However, to test the model and run it in real time, we have deployed the model as a Web App (refer section IV.B.1). Tables I II III IV V demonstrate some of the predictions from the test dataset.

| Language Pair | Input Word | Predicted Word |
|---|---|---|
| Eng-Hin | MUSEUM | म्यूज़ियम |
| Eng-Hin | UNIVERSITY | यूनिवर्सिटी |
| Eng-Ben | Sahajadibibi | সাহাজাদিবিবি |
| Eng-Ben | Antarctica | আন্টকটিকা |
| Eng-Tam | CORPORATION | கார்ப்பரேஷன் |
| Eng-Tam | PHARMACEUTICAL | பார்மாசெட்டிக்கல் |
| Eng-Kan | GOVERNOR | ಗವರ್ನರ್ |
| Eng-Kan | MERIT | ಮೆರಟ್ |

TABLE I: Few examples of transliterated outputs from the model, for all language pairs with source as English.

| Language Pair | Input Word | Predicted Word |
|---|---|---|
| Hin-Eng | क्वीन्सलैंड | QUEENSLAND |
| Hin-Eng | मॅटोग्रोसेंस | MATOGROSSENSE |
| Hin-Ben | अनवर | আনওয়ার |
| Hin-Ben | गोडल | গোডেল |
| Hin-Tam | आसही | அசாஹி |
| Hin-Tam | केनन | கென்னன் |
| Hin-Kan | मुईज़ | ಮುಯೀಜ್ |
| Hin-Kan | आसिफ | ಆಸಿಫ್ |

TABLE II: Few examples of transliterated outputs from the model, for all language pairs with source as Hindi.

| Language Pair | Input Word | Predicted Word |
|---|---|---|
| Ben-Eng | নমিতারাণী | Namitarani |
| Ben-Eng | সালাহুদ্দিন | Salahuddin |
| Ben-Hin | রামফল | रामफल |
| Ben-Hin | ধরমবীর | धरमबीर |
| Ben-Tam | হোমে | ஹோம்மே |
| Ben-Tam | মৌসম | மௌசம் |
| Ben-Kan | অলকা | ಅಲೆಂಕ |
| Ben-Kan | অলীক | ಅಲೀಕ್ |

TABLE III: Few examples of transliterated outputs from the model, for all language pairs with source as Bengali.

| Language Pair | Input Word | Predicted Word |
|---|---|---|
| Tam-Eng | மட்டக்ரோசென்ஸ் | MATOGROSSENSE |
| Tam-Eng | ராஜகன்வாரீ | RAAJAKUNVAREE |
| Tam-Hin | தைபே | तायपेई |
| Tam-Hin | கஹானி | कहानी |
| Tam-Ben | கரலே | কারালে |
| Tam-Ben | அசோக | অশোক |
| Tam-Kan | ஹூயே | ಹುಯ್ |
| Tam-Kan | அஸோம் | ಅಸೋಮ್ |

TABLE IV: Few examples of transliterated outputs from the model, for all language pairs with source as Tamil.

.

| Language Pair | Input Word | Predicted Word |
|---|---|---|
| Kan-Eng | ಪ್ಯಾರೆಮೋಹನ್ | PYAREMOHAN |
| Kan-Eng | ವಿಜಿತೇಂದ್ರಿಯ | VIJITENDRIY |
| Kan-Hin | ಲಡಕಿ | लड़की |
| Kan-Hin | ಕಿರಣ | किरण |
| Kan-Ben | ಅಂಗರ | আঙ্গারা |
| Kan-Ben | ಸಫರ್ | সাফার |
| Kan-Tam | ಅಚಲಾ | அச்சலா |
| Kan-Tam | ಆಲಂ | ஆல்ஹா |

TABLE V: Few examples of transliterated outputs from the model, for all language pairs with source as Kannada.

*1) Web API:* We have deployed MATra as an API on CellStrat Hub and created a web app for easy access to the model. However, we have rate-limited the requests and this application should only be used for experiments and not for production use cases. Anyone can access the model's web app at https://bi-matra.netlify.app/

To use the app, user must input their word in the source language, and select their input and target languages respectively to get the transliterated text. We have also added the feature of transliterating sentences in the website. Figures 5 and 6 show the screenshots of the results from the web app.

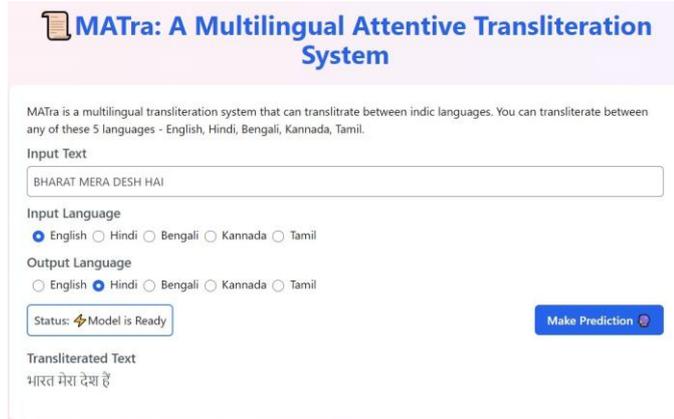

Fig. 5: Sample output for the sentence 'BHARAT MERA DESH HAI' from English to Hindi.

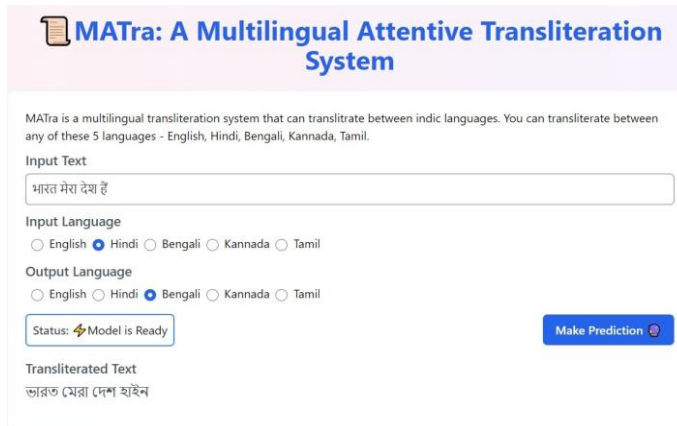

Fig. 6: Sample output of the same sentence transliterated from Hindi to Bengali.

## V. RESULTS

As mentioned earlier, there is no direct dataset available for training the model, so we built a bi-directional multilingual model that uses English as an intermediate for transliterating between Indic languages. Due to the lack of a dataset in both training and testing, automated evaluations aren't possible, instead, we do Human Evaluation of the results using crowd-sourced evaluators. The primary results for our task (Indic to Indic transliteration) are completely based on human evaluators and beat the state of art transliteration models that are released publicly. Evaluators evaluate the model predictions based on the phonetic similarity of the predicted words with the input words. They are also instructed to write down the correct transliterated word if the model's prediction isn't satisfactory. This helps us compute some other metrics like CER and BLEU scores as well apart from the usual accuracy.

As our model is based on chaining the results of the Indic to English and English to Indic tasks, we evaluate our model on these foundational tasks as well, because we have no other metrics to compare with the existing models. All other papers in this domain have only published their top-1 accuracy scores, and that is the only available benchmark we can compare our model with. Our model has achieved significantly better results in terms of accuracy and is able to beat the state-of-the-art models by a large margin (further discussed in section V.B.I).

**However, the task of transliteration is a phonetic/sound-based task, and the metrics that must be used to evaluate this model should also be phonetic/sound based or should either represent a score that can show the phonetic errors. All other papers that are relevant to ours have used only the top-1 accuracy score, which might be a good metric for evaluation, but it is not the ideal metric to evaluate this task**. We have also compared those models using top-1 accuracy (section V.B.I) because we have no other metric for comparison. **However, we use Phonetic_Accuracy as our primary metric of evaluation**.

### A. Indic to Indic Transliteration

*1) Phonetic Accuracy:* Transliteration is a sound-based task, and thus, we use Phonetic Accuracy [22] as the primary metric to evaluate our model. These scores are generated upon calculating how correct each prediction is based on the phonetic sound. These scores are human evaluated, and thus we can notice that in the real world, after testing the results by humans, the model is extremely accurate in almost every case.

Formula (1) shows the formula for the phonetic accuracy. This metric is basically similar to the accuracy scores seen on general tasks, but here it is human evaluation instead of computer evaluation, and criterion is phonetic similarity instead of correctness.

$$Accuracy_{Phonetic} = \frac{Correct\ sounding\ words}{Total\ number\ of\ words} \quad (1)$$

Table VI shows the Phonetic Accuracy of the model between the different Indic pairs. Average Phonetic Accuracy across all Indic language pairs is 0.935.

**In general, each language pair is able to easily reach an accuracy score of about 90% at least, with the highest going to about 97%. For Indic to Indic transliteration**

|        | Hindi | Bengali | Tamil | Kannada |
|--------|-------|---------|-------|---------|
| Hindi  | -     | 0.937   | 0.968 | 0.935   |
| Bengali| 0.946 | -       | 0.922 | 0.905   |
| Tamil  | 0.913 | 0.939   | -     | 0.942   |
| Kannada| 0.948 | 0.932   | 0.944 | -       |

TABLE VI: Phonetic Accuracy (higher the better). *The row indicates the source language and the column indicates the target language.*

**tasks, the results generated by our model are superior to all existing models/approaches available and predictions are better and highly precise in comparison. It must be noted that achieving scores like 90% or higher is a really challenging task in this domain.**

*2) CER (Character Error Rate) :* Character Error Rate [23] is a measure of how far off the predicted word is from the ground truth word at a character level. CER basically tells us the minimum number of character level operations (substitutions, deletions or insertions) required to obtain the ground truth word from the predicted word. The more the character operations required, the more the error rate and lesser the performance of the system. Phonetically speaking, slight changes in the characters still sound the same and that's why we also use the CER metric to evaluate our model.

Vanilla CER between a predicted word and the ground truth is calculated as shown below:

$$CER = \frac{S + D + I}{N} \quad (2)$$

Where,
S = Number of Substitutions
D = Number of Deletions
I = Number of Insertions
N = Number of Characters in Ground Truth

However, with this formulation, the CER values can exceed 1 (100%) especially in cases where many insertions are required. To get around this issue and make sure the CER can be expressed as a percentage, the CER is normalized as follows:

$$CER_{normalized} = \frac{S + D + I}{S + D + I + C} \quad (3)$$

Where additionally, C = Number of Correct Characters

The CER scores are calculated by testing on the NEWS 2018 test datasets which were annotated by human evaluators. All other papers in this domain relevant to our task have not published their CER scores, so we do not have any benchmarks to compare in terms of this metric.

Table VII shows the CER results between the different Indic pairs. Average CER across all Indic language pairs is 0.071.

|        | Hindi | Bengali | Tamil | Kannada |
|--------|-------|---------|-------|---------|
| Hindi  | -     | 0.058   | 0.085 | 0.073   |
| Bengali| 0.053 | -       | 0.100 | 0.080   |
| Tamil  | 0.057 | 0.079   | -     | 0.081   |
| Kannada| 0.036 | 0.064   | 0.085 | -       |

TABLE VII: Character Error Rate (lower the better). *The row indicates the source language and the column indicates the target language.*

*3) BLEU:* The Bilingual Evaluation Understudy [24] (called as BLEU in short) is a metric used for evaluating the predictions done by NLP models for any sequence-to-sequence task. It basically shows the fraction of tokens or words covered by the predicted sentence as compared to the real/truth sentences. For our task, we have letters instead of real words and sentences, so we scored each prediction on character level instead of word level (which is done on most translation systems). The range of scores can vary between 0 to 1, showing how good does the model perform for each prediction (higher the better). If the score is close to 1, it means the word is perfectly correct and each letter in the word is part of the real/truth word in the correct order.

The calculation of this score requires 2 inputs, a predicted sequence and the real/truth sequence. The predicted sequence is anyway generated by our model, but in case of the truth/real sequence, they were given by human evaluators. For each word in our test dataset, the human evaluator writes down the correct transliterated word (if prediction is incorrect) and the BLEU score is then calculated with reference to the word mentioned by the evaluator.

Tables VIII IX X XI show the individual and cumulative BLEU 1, 2, 3 and 4 scores respectively. Related work in this domain don't have published BLEU scores for their models so we do not have any benchmark to compare and beat in terms of this metric.

| BLEU-1 Scores | | | | |
|---|---|---|---|---|
|         | Hindi | Bengali | Tamil | Kannada |
| Hindi   | -     | 0.96    | 0.93  | 0.94    |
| Bengali | 0.95  | -       | 0.92  | 0.93    |
| Tamil   | 0.92  | 0.92    | -     | 0.91    |
| Kannada | 0.97  | 0.96    | 0.93  | -       |

TABLE VIII: Pairwise BLEU-1 Scores

| BLEU-2 Individual / Cumulative Scores | | | | |
|---|---|---|---|---|
|         | Hindi     | Bengali   | Tamil     | Kannada   |
| Hindi   | -         | 0.91 / 0.93 | 0.84 / 0.87 | 0.87 / 0.90 |
| Bengali | 0.90 / 0.92 | -         | 0.81 / 0.86 | 0.84 / 0.88 |
| Tamil   | 0.88 / 0.89 | 0.86 / 0.88 | -         | 0.84 / 0.86 |
| Kannada | 0.93 / 0.95 | 0.90 / 0.92 | 0.84 / 0.88 | -         |

TABLE IX: Pairwise BLEU-2 Scores

| BLEU-3 Individual / Cumulative Scores | | | | |
|---|---|---|---|---|
|         | Hindi     | Bengali   | Tamil     | Kannada   |
| Hindi   | -         | 0.89 / 0.91 | 0.81 / 0.84 | 0.85 / 0.87 |
| Bengali | 0.88 / 0.90 | -         | 0.78 / 0.82 | 0.80 / 0.84 |
| Tamil   | 0.91 / 0.89 | 0.88 / 0.87 | -         | 0.86 / 0.87 |
| Kannada | 0.92 / 0.93 | 0.87 / 0.90 | 0.81 / 0.85 | -         |

TABLE X: Pairwise BLEU-3 Scores

| BLEU-4 Individual / Cumulative Scores | | | | |
|---|---|---|---|---|
|         | Hindi     | Bengali   | Tamil     | Kannada   |
| Hindi   | -         | 0.93 / 0.91 | 0.87 / 0.84 | 0.90 / 0.87 |
| Bengali | 0.94 / 0.90 | -         | 0.83 / 0.81 | 0.85 / 0.83 |
| Tamil   | 0.96 / 0.90 | 0.94 / 0.88 | -         | 0.91 / 0.86 |
| Kannada | 0.95 / 0.93 | 0.90 / 0.89 | 0.83 / 0.83 | -         |

TABLE XI: Pairwise BLEU-4 Scores

*B. Indic to English and English to Indic Transliteration*

Our model uses the English transliterated word as the intermediate output for this task, and thus, in our approach,

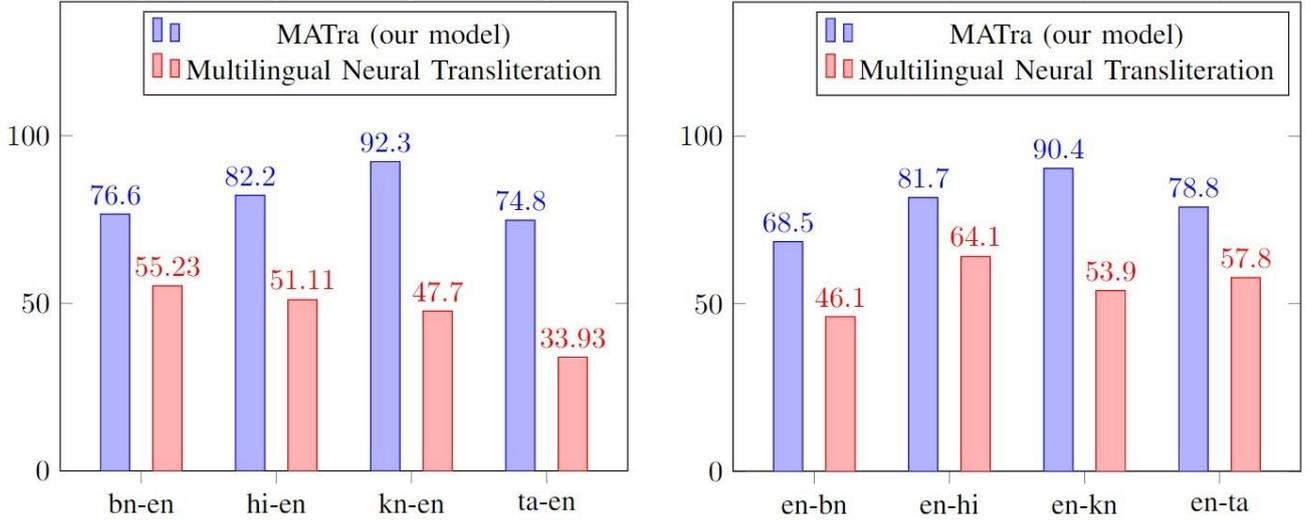

Fig. 7: Multilingual Neural Transliteration refers to the paper Leveraging Orthographic Similarity for Multilingual Neural Transliteration [6]. This figure shows the top-1 accuracy for the task of transliteration for all 4 Indic languages (Bengali, Hindi, Kannada, Tamil) to English and English to all 4 Indic languages (Bengali, Hindi, Kannada, Tamil) for both papers respectively.

the Indic-to-Indic transliteration task is derived from two foundational tasks: Indic to English and English to Indic. **Our primary task is transliterating from Indic-to-Indic languages, and the actual metric of evaluation is Phonetic Accuracy (section V.A.1) which is a phonetic based metric. However, no other paper has used any phonetic based evaluation system, so we are compelled to compare results based on metrics like Top-1 accuracy**.

*1) Top-1 Accuracy:* In our experiments, we find that our Bi-Directional model (trained on both tasks together) outperforms the Uni-Directional baselines (trained on each task independently). Fig 8 shows the Absolute Accuracy (*this is the computer-evaluated accuracy based on the test dataset and is NOT based on phonetics like the human evaluated results*) of Uni-Directional and Bi-Directional models on the two tasks.

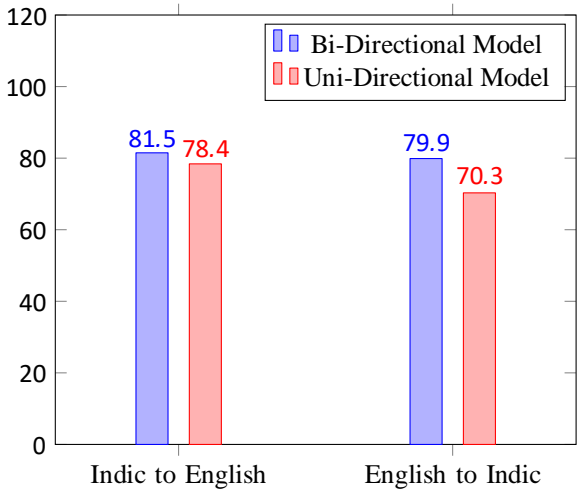

Fig. 8: Absolute Accuracy (higher the better) of the two types of models on the Indic to English and English to Indic foundational tasks. Bi-Directional model clearly outperforms the Uni-Directional model on both the tasks. On Indic to English, uni-directional model achieves an absolute accuracy of 78.4% while the bi-directional model achieves an accuracy of 81.5%. On English to Indic, the uni-directional model achieves an accuracy of 70.3% while the bi-directional model achieves an accuracy of 79.9%, with a combined average accuracy score of 80.7%.

The bi-directional model is clearly the winner in terms of accuracy and we believe this is due to the dataset being twice the size of the dataset of the uni-directional ones and validates the notion that transformers work better with larger datasets.

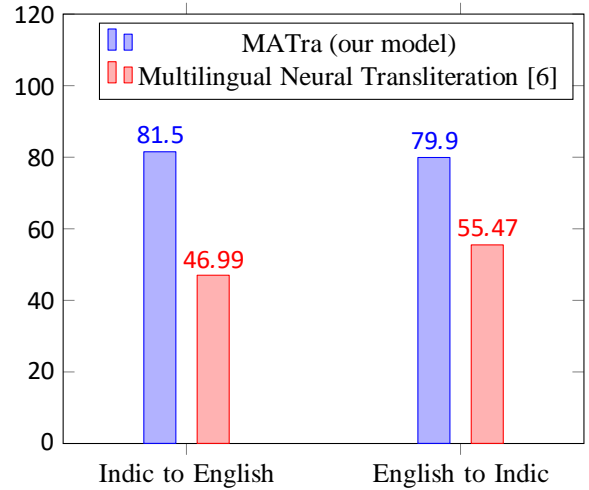

Fig. 9: Multilingual Neural Transliteration refers to the paper Leveraging Orthographic Similarity for Multilingual Neural Transliteration [6]. This figure shows the top-1 accuracy scores of both models for the task of Indic to English and English to Indic Transliteration for both papers.

We also pseudo-compare our results with other papers in terms of absolute accuracy. However, the other papers haven't released the dataset publicly for us to evaluate our model on a standard benchmark. So, the comparative scores are independent of the dataset and are just raw number comparisons.

The first paper (Leveraging Orthographic Similarity for Multilingual Neural Transliteration [6]) claims a score of 51.23% at average on their own Indic-Eng and Eng-Indic dataset. Fig 9 shows the scores achieved by this paper in comparison to our scores.

The authors of Brahmi-Net [7] have published scores for 4 language pairs (ben-mar, mal-tam, mar-ben and tel-mar) with an average score of 59.35%. They have performed their evaluation on a private dataset which is not accessible by us for comparison. Furthermore, they have not published scores for any language pair that matches our task (though they have built their system on all pairs that we are working on), so we could not compare their model with ours.

Fig 7 shows the top-1 accuracy scores for all the language pairs that involve English (in either input or output). No scores of Brahmi-Net [7] are considered as the authors have not mentioned the related scores in their paper. The only comparison that can be done is on the paper Leveraging Orthographic Similarity for Multilingual Neural Transliteration [6] and **there is a minimum increase of 20% on all language pairs** (Eng-Hin is an exception, with only 17.6% increase).

These were the results of our model in comparison to earlier models. **The primary task of our model is Indic to Indic transliteration (not Indic to Eng or Eng to Indic), and these metrics are just to represent the improvements over earlier models/approaches. Furthermore, Top-1 accuracy was the only metric available for comparison, so in order to show our improvements, it was mandatory to do the required comparisons. However, the real metric to represent our model performance is Phonetic_Accuracy, which is elaborated in section V.A.1**.

Accuracy is a good metric for evaluating general tasks where the requirement is to check for the correct prediction. However, transliteration is a phonetic based task and the ideal metric for evaluation must be able to convey how accurate is the performance of the model on the basis of sounds/phonetics. Considering all the acceptable ambiguities in such tasks, it should be evident that raw accuracy scores are clearly insufficient to completely evaluate any model in such tasks at least.

*2) CER:* Fig 15 shows the Character Error Rate (CER) of Uni-Directional and Bi-Directional models on the two tasks.

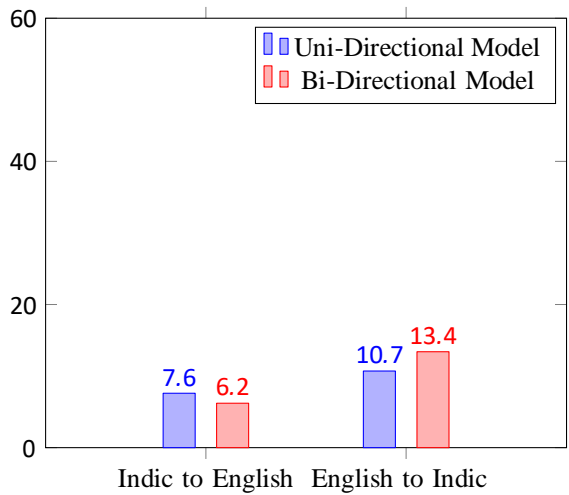

Fig. 10: Character Error Rate (lower the better) of the two types of models on the Indic to English and English to Indic foundational tasks. On Indic to English, the uni-directional model has a CER of 7.6% while the bi-directional model has a better CER of 6.2%. However, on English to Indic, the uni-directional model has a CER of 10.7% while the bi-directional model has a worse CER of 13.4% which is unexpected given the fact that the accuracy is better than uni-directional models for both tasks.

In terms of CER, the bi-directional model has a relatively worse CER than the uni-directional ones in the English to Indic task. The reason behind this discrepancy between the accuracy being better and the CER being worse is because of the fact that the words that were predicted wrong by the bi-directional model had higher errors in the characters compared to the ground truth words.

## VI. LIMITATIONS

Transformers have proven to excel in performance at scale. Here the size of the dataset is relatively small and hence the absolute accuracy still has room for improvement even though phonetic accuracy is good. Apart from room from improved accuracy, the current inference architecture is a two stage process of first transliterating between Indic to English and then feeding this English output as the input back to the model for transliterating to the target Indic language. This makes the inference time twice than what it would have taken if it was an end-to-end model for direct transliteration from Indic to Indic.

## VII. FUTURE WORK

Transliterating between Indic Languages is challenging due to the lack of public datasets on the task. We tackle this problem by using English as the intermediate representation i.e., chaining together Indic to English and English to Indic in two stages. This requires performing inference twice using the same model for one pair. Our next steps would be to create an End-to-End model that doesn't need to be chained together in two stages and eliminate the need of generating the English intermediate and directly transliterate between Indic pairs while not actually having the dataset to train such a model as mentioned earlier. We intend to experiment with knowledge distillation techniques to achieve this by using our current bi-directional model as the teacher.

## VIII. CONCLUSION

Our model is based on the Vanilla Transformers architecture with modifications to the training procedure using special language tokens for multilingual transliteration, and it is able to beat the state-of-the-art models in this domain by a large margin. We perform the task of Indic-to-Indic transliteration without a direct dataset. **We achieve a score of 80.7%, outperforming all other models by about 29.5% in terms of absolute accuracy and achieve an average Phonetic Accuracy of 93.5%**.

### ACKNOWLEDGMENT

The authors wish to acknowledge the help and support provided by the open-source community for their contributions and Dr. Sachi Pandey, Assistant Professor at SRM Institute of Science and Technology, for her supervision and feedback on the paper. They would also like to thank the evaluators for helping in evaluating the respective models.